\documentclass{article}
\usepackage{spconf,amsmath,graphicx}
\usepackage{hyperref}
\usepackage{times}
\usepackage{epsfig}
\usepackage{graphicx}
\usepackage{amsmath}
\usepackage{amssymb}
\usepackage{algorithm} 
\usepackage{algpseudocode}  % Algorithm

\usepackage[square,numbers]{natbib}
\usepackage{times,epsfig,graphicx,amsmath,amssymb}
\usepackage{times,graphicx,amsmath,amssymb,booktabs,tabulary,multirow,overpic,bbm,amssymb}

\newcolumntype{x}[1]{>{\centering\arraybackslash}p{#1pt}}

\newlength\savewidth\newcommand\shline{\noalign{\global\savewidth\arrayrulewidth
  \global\arrayrulewidth 1pt}\hline\noalign{\global\arrayrulewidth\savewidth}}

\makeatletter\renewcommand\paragraph{\@startsection{paragraph}{4}{\z@}
  {.5em \@plus1ex \@minus.2ex}{-.5em}{\normalfont\normalsize\bfseries}}\makeatother

\title{LOOC: Localize Overlapping Objects with Count Supervision}
\name{Issam H. Laradji$^{1,2}$, Rafael Pardinas$^{1}$, Pau Rodriguez$^{1}$, David Vazquez$^{1}$}
\address{$^{1}$Element AI, Montreal, Canada \{issam.laradji, rafael.pardinas, pau.rodriguez, dvazquez\}@elementai.com\\ $^{2}$ University of British Columbia}
\begin{document}
%\ninept

\maketitle

\begin{abstract}
Acquiring count annotations generally requires less human effort than point-level and bounding box annotations. Thus, we propose the novel problem setup of localizing objects in dense scenes under this weaker supervision. We propose LOOC, a method to Localize Overlapping Objects with Count supervision.  We train LOOC by alternating between two stages. In the first stage, LOOC learns to generate pseudo point-level annotations in a semi-supervised manner. In the second stage, LOOC uses a fully-supervised localization method that trains on these pseudo labels. The localization method is used to progressively improve the quality of the pseudo labels. We conducted experiments on popular counting datasets. For localization, LOOC achieves a strong new baseline in the novel problem setup where only count supervision is available. For counting, LOOC outperforms current state-of-the-art methods that only use count as their supervision. Code is available at:~\url{https://github.com/ElementAI/looc}.
\end{abstract}


%%%%%%%%%%%%%%%%%%%%%%%%%%%%%%%%%%
% INTRO
%%%%%%%%%%%%%%%%%%%%%%%%%%%%%%%%%%
\section{Introduction}

% Context and motivation
We consider the task where a model has to predict the location of each object in an image. This task is important for applications such as public safety, crowd monitoring, and traffic management. Typically, bounding boxes~\cite{ren2015faster, redmon2016you} or point-level annotations~\cite{laradji2018blobs, laradji2019instance, li2018csrnet, lempitsky2010learning, zhang2016single, onoro2016towards, cholakkal2019object} are provided during training. However, we consider the more challenging problem setup where only count-level annotations are available. These labels are cheaper to acquire than point-level annotations, but they make the localization task significantly more difficult for the model. In dense scenes, the model has to identify which objects in the image correspond to the object count. These objects can heavily overlap, can vary widely in scale, shape, and appearance. Current methods~\cite{Gao2018cwsl} partially address this problem setup but only for datasets where objects are salient and rarely overlap. These methods do not work for dense scene datasets as they are designed to work with training images that have few objects. Thus, we address a novel problem setup of learning to localize objects for dense scenes under count supervision.

% Why count supervision
Acquiring object count labels in images requires much less human effort than annotating the location of each object. For training images with $4$ or less objects, the annotator can obtain the object count much faster than with point annotations through subitizing~\cite{chattopadhyay2016counting}. For videos, the annotator can obtain the object count quickly across image frames as the count changes much less frequently than the object locations in the video. In some cases, object counts can be obtained with no effort compared to point-level annotations. These cases include keeping count of products on retail stock shelves, and keeping count of a crowd of people at events where the ticket system registers their actual count. In both cases, identifying object locations is important for safety and logistics.

\begin{figure}[t]
  \centering
  \includegraphics[trim={2cm 1.9cm 2cm 3.2cm},clip, width=\linewidth]{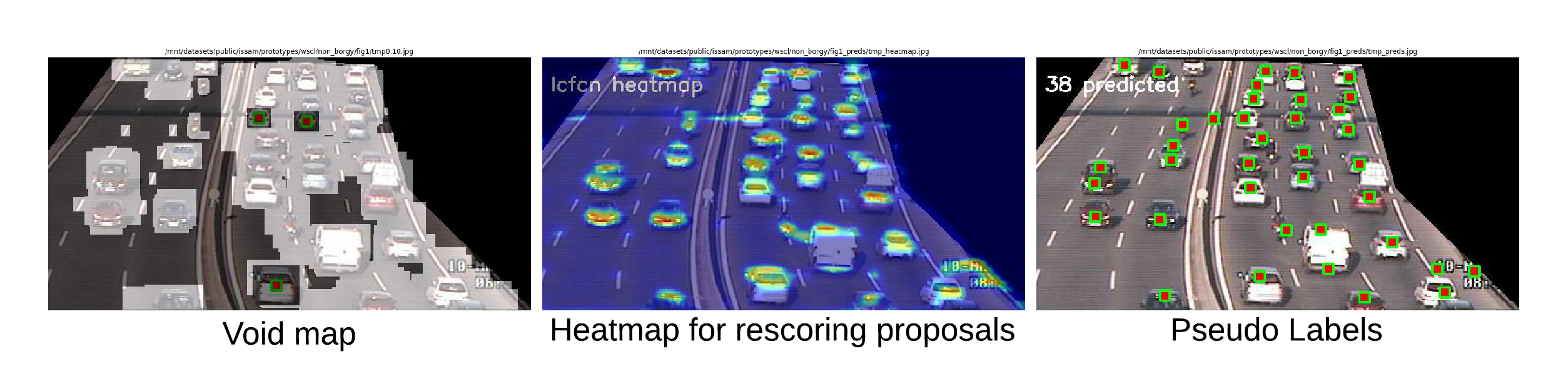}
  \caption{\textbf{Generating pseudo labels.} (left) we use the centroids of the highest scoring proposals as initial labels, and consider the lower scoring proposals as unlabeled (shown bright). (middle) the model trains on these initial labels and infers the pixels that belong to objects in the unlabeled region. (right) using the trained model, we re-score the proposals to obtain more confident pseudo labels.}
  \label{fig:PN}
\end{figure}

% Point level annotations (LC-FCN and CSRNet)
Many methods exist that can perform object localization but they need to be trained on point-level annotations~\cite{lempitsky2010learning, laradji2018blobs, li2018csrnet, laradji2019instance} or image-level~\cite{laradji2019masks}. They fall under two main categories: density-based and segmentation-based localization. Density-based methods~\cite{lempitsky2010learning, li2018csrnet} transform the point-level annotations into a density map using a Gaussian kernel. Then, they train using a least-squares objective to predict the density map. However, these methods do not provide individual locations of the objects. On the other hand, segmentation-based methods such as LC-FCN~\cite{laradji2018blobs} train using a loss function that encourages the output to contain a single blob per object. For our framework, we use the individual object locations obtained by LC-FCN to help in generating the pseudo point-level annotations.

% Methods using count supervision
Glance~\cite{chattopadhyay2016counting} is a state-of-the-art method for object counting when only count supervision is provided. This method is an ImageNet~\cite{deng2009imagenet} pre-trained model such as ResNet50~\cite{he2016deep} with a regression layer as its output layer. Unfortunately, Glance is not designed to localize the objects of interest in the image. In this work, we propose a novel approach that uses count supervision to localize objects in dense scenes. Further, we show that our method achieves better count results than Glance.

\begin{figure*}[t]
  \centering
  \includegraphics[trim={0 3cm 0 3cm},clip,width=0.9\linewidth]{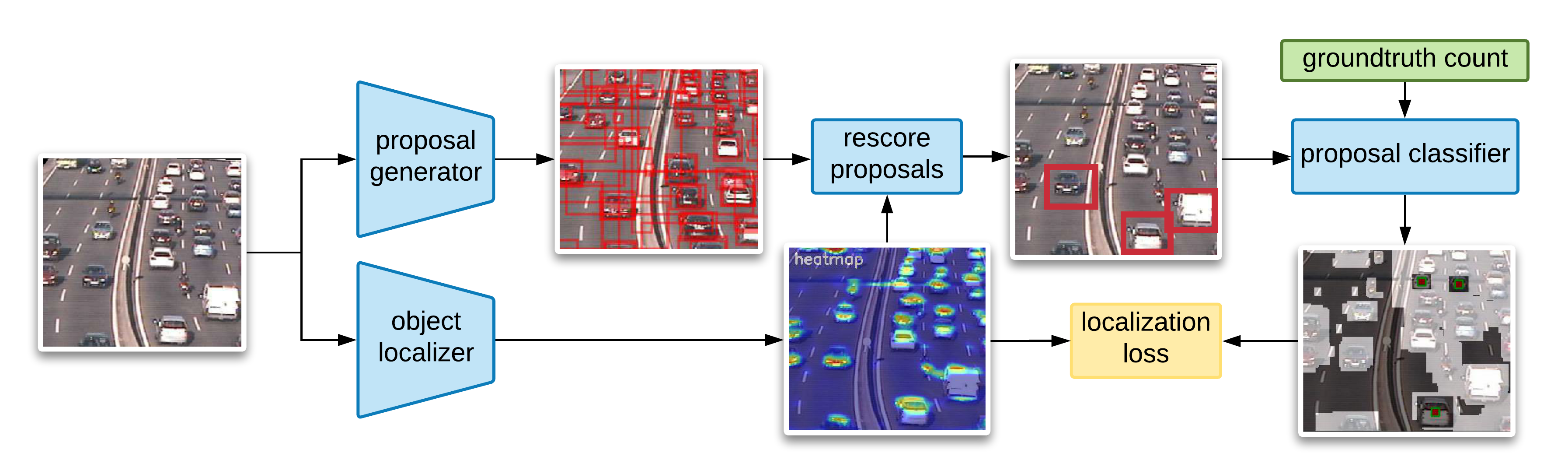}
  \caption{\textbf{Training.} LOOC is composed of two branches. The first one is a pseudo label generation branch that obtains point-level labels by selecting the highest scoring object proposals. The second one is a localization branch that trains a fully supervised localization model on the pseudo labels. This branch is also used to re-score the object proposals.}
  \label{fig:looc}
\end{figure*}

% Weakly supervised localization methods
Most weakly supervised localization methods fall under multiple-instance learning (MIL)~\cite{MIL}. In this setup, each image corresponds to a bag of object proposals. Each bag is labeled based on whether an object class exists. \citet{Li2016WeaklySO} present a two-step approach. First, they use a mask-out strategy to filter the noisy object proposals; then, they use a Faster RCNN~\cite{ren2015faster} for detection using bags of instances.  \citet{tang2017multiple} use a refinement learning strategy to improve on the quality of the proposals. C-MIL~\cite{Wan2019CMILCM} introduces a continuation optimization method to avoid getting stuck in a local minima. C-WSL~\cite{Gao2018cwsl} is the most relevant to our work as they use count information to obtain the highest scoring proposals. However, it differs from our setup in that it relies on a classification network that is not designed for dense scenes.

% Our method
We propose LOOC which can learn to Localize Overlapping Objects with Count supervision. It trains by alternating between two stages. In the first stage, LOOC learns to generate pseudo point-level annotations in a semi-supervised learning manner. In the second stage, LOOC uses a fully-supervised localization method that trains on these pseudo labels. 

As in Figure~\ref{fig:PN}, the pseudo labels are the centroids of the highest scoring proposals generated using a standard proposal method such as selective search~\cite{uijlings2013selective}. This set of scores is the combination of the proposal objectness and the probability heat-map obtained from the trained localization method. The proposals that have low scores are considered unlabeled. The localization method uses the pseudo labels and ignores the regions that are unlabeled. The goal for the localization method is to infer the object probabilities in these unlabeled regions. These probabilities are used to re-score the proposals to generate the pseudo labels in the next round~\ref{fig:looc}. At test time, only the localization method is kept, which can be directly used to predict the locations and count of the objects of interest.

% Results
Since no direct relevant work exists for this particular setup, we compare our methods against Glance~\cite{chattopadhyay2016counting} and the fully supervised LCFCN~\cite{laradji2018blobs}. We benchmark our methods against various counting datasets such as Trancos~\cite{guerrero2015Trancos}, Penguins~\cite{arteta2016counting}, UCSD~\cite{chan2008privacy}, and Mall~\cite{chen2012feature}. We observed that LOOC achieves a strong new baseline in the novel problem setup where only count supervision is available with respect to localization. Further, we observed that LOOC outperforms current state-of-the-art methods that only use count as their supervision. 

% Summary of contributions
We summarize our contributions as follows: we (1) present LOOC, a novel framework that can count and locate objects with count-level supervision for dense scenes; (2) propose a semi-supervised learning scheme where pseudo labels are inferred for unlabeled regions in the image, and (3) show that LOOC achieves better count accuracy than Glance with the addition that it locates objects efficiently.

%%%%%%%%%%%%%%%%%%%%%%%%%%%%%%%%%%
% METHOD
%%%%%%%%%%%%%%%%%%%%%%%%%%%%%%%%%%
\section{Proposed Method}

\subsection{Motivation} 
One of the main challenges of training with only count supervision is to identify which objects of interest in the image correspond to the object count. Object proposals could be used to identify which regions are likely to have the objects of interest~\cite{Tang2018ECCV, Bilen2016CVPR, Zhou2018PRM}. However, proposal methods are class-agnostic as they do not provide the class label. Thus, they might propose the wrong objects.

To alleviate this drawback, we consider a semi-supervised learning methodology where only the centroids of the proposals with the highest saliency score are considered as pseudo point-level annotations. The rest of the proposals represent unlabeled regions. When a localization model is trained on these salient proposals, it can be used to predict a class probability map (CPM) for the objects of interest that are in the unlabeled regions. These probabilities are used as positive feedback to re-score the proposals and obtain better pseudo point-level annotations for the next round.

\subsection{Framework}
Figure~\ref{fig:looc} illustrates the pipeline of our framework LOOC. It consists of three components: a proposal generator, a proposal classifier, and an object localizer. The proposal generator and classifier are used to obtain the pseudo point-level annotations, whereas the object localizer is trained on these annotations to count and localize objects. We explain each of these components below.

\subsection{Generating pseudo-labels}
In this section, we explain the proposal generator and the classifier and how they can be used to generate pseudo point-level annotations. 

First, a proposal generator such as selective search~\cite{uijlings2013selective} is used to output 1000 proposals that correspond to different objects in the image. Each of these proposals has an associated score obtained from the object localizer (see Section~\ref{sec:localizer} for more detail). The proposal classifier uses these scores to obtain labeled and unlabeled regions in the training images. The regions that do not intersect with any proposal are labeled as background whereas the region that intersect with the $r$ highest-scoring proposals are labeled as foreground. The remaining regions are considered unlabeled.

The highest scoring proposals are selected using non-maximum suppression~\cite{neubeck2006efficient}, and their centroids are considered as the pseudo point-level annotations used to train the object localizer.

\subsection{Training a Localization Method}
\label{sec:localizer}
Using the pseudo point-level labels, we can train any fully supervised localization network such as LC-FCN~\cite{laradji2018blobs} and CSRNet~\cite{li2018csrnet}. However, we chose LC-FCN due to its ability to get a location for each object instance rather than a density map. For the point annotations in the labeled regions, LC-FCN is trained using its original loss function described in detail by~\citet{laradji2018blobs}. 

LC-FCN's predictions on the unlabeled regions are ignored during training. However, the class probability map (CPM) that LC-FCN outputs for those regions is used to re-score the proposals in order to obtain a new set of pseudo point-level annotations.

\subsection{Overall Pipeline}

% the predictions made by LC-FCN are ignored with respect to the loss function. As a result, we anticipate that LC-FCN interpolates the locations of the objects of interest in the unlabeled regions. This interpolation is used as positive feedback by the proposal classifier to re-score the proposals and obtain new pseudo point-level annotations.

LOOC is trained in cycles where in each cycle it alternates between generating pseudo point-level annotations and training LC-FCN on those labels (Algorithm~\ref{alg:TheAlgorithm}). Let $c_i$ be the true object count for image $i$. At a given cycle, we only consider the top $r_i$ scoring proposals (where $r_i\leq c_i$) to be used for obtaining the pseudo point-level annotations. After training LC-FCN with the top $r_i$ proposals, we use its class probability map (CPM) to re-score the proposals and increase $r_i$ by $\delta$.\footnote{we used $r_i=0.1$ and $\delta=0.1$} The score of each proposal is the mean of CPM's region that intersects with that proposal. This allows us to pick a larger number of pseudo point-level annotations and increase the size of the labeled regions. The procedure ends when $r_i$ equals $c_i$ for all images, which closely resembles curriculum learning~\cite{Bengio2009CurriculumL} under an expectation maximization framework~\cite{Altri1977MaximumLF}. 

%%% Alg: LOOC training %%%
\begin{algorithm}[t]
  \caption{LOOC training}
  %\resizebox{0.8\columnwidth}{!}{%
  \begin{algorithmic}[1]
    \State $r := 0.1$
    \While {$r <= 1$} 
    \State Obtain LC-FCN's class probability map (CPM)
      \State Generate proposals $P$
      \State Compute scores $S_{P}$ using CPM 
      \State Select top $r \cdot c$ proposals 
        {\small (c is the object count)}
      \State Obtain labeled and unlabeled regions for all images
        \State Train LC-FCN on the labeled regions
      \State $r = r + \delta$ (increase ratio of selected proposals)
    \EndWhile
    \State Generate the final pseudo point-level annotations
    \State Train LC-FCN on these labels
  \end{algorithmic}
  %}
\label{alg:TheAlgorithm}
\end{algorithm}

%%%%%%%%%%%%%%%%%%%%%%%%%%%%%%%%%%
% EXPERIMENTAL SETUP
%%%%%%%%%%%%%%%%%%%%%%%%%%%%%%%%%%
\section{Experiments}

\begin{table*}[t]
    \centering
    %\resizebox{\textwidth}{!}{%
    \begin{tabular}{l|x{20}|x{28}|x{20}|x{28}|x{20}|x{28}|x{20}|x{28}|x{20}|x{28}|}
        \multicolumn{1}{c|}{\textbf{}} &
        \multicolumn{2}{c|}{\textbf{USCD}} &
        \multicolumn{2}{c|}{\textbf{Trancos}} &
        \multicolumn{2}{c|}{\textbf{Mall}} &
        %\multicolumn{2}{c|}{\textbf{WorldExpo}} &
        \multicolumn{2}{c|}{\textbf{Penguins}}\\
        & MAE & GAME & MAE & GAME & MAE & GAME & MAE & GAME \\\shline
        Glance~\cite{chattopadhyay2016counting} & 3.80 & -    & 11.66 & -     & 4.67 & -    & 14.79 & - \\
        LC-FCN~\cite{laradji2018blobs}          & 1.57 & 4.35 & 5.50  & 9.73  & 2.64 & 7.26 & 7.59  & 13.01\\
        % Dense~\cite{} & & & & & & & & \\\hline
        TopK (Ours)                             & 2.41 & 6.85 & 9.06 & 15.74 & 3.49 & 16.78 & 8.85  & 17.53\\
        LOOC (Ours)                             & 2.20 & 6.74 & 8.68 & 14.90 & 3.23 & 16.51 & 8.42  & 15.97 \\
    \end{tabular}
    %}
    %\vspace{3mm}
    \caption{\textbf{Count and localization results across dense scene validation set.}}
    \label{tab:results}
\end{table*}

\begin{table}[t]
    \centering
    \resizebox{\linewidth}{!}{%
    \begin{tabular}{l|x{28}|x{28}|x{28}|x{28}|}
        \multicolumn{1}{c|}{\textbf{}} &
        \multicolumn{1}{c|}{\textbf{USCD}} &
        \multicolumn{1}{c|}{\textbf{Trancos}} &
        \multicolumn{1}{c|}{\textbf{Mall}} &
        \multicolumn{1}{c|}{\textbf{Penguins}}\\\shline
        TopK (Ours) & 7.29 & 12.27 & 15.47 & 6.92\\
        LOOC (Ours) & 6.02 & 10.46 & 14.76 & 5.74\\
    \end{tabular}
    }
    \caption{\textbf{Ablations studies}. Localization results on the training set to evaluate the quality of the generated pseudo point-level annotations.}
    \label{tab:ablation_results}
\end{table}

% Datasets and metrics
In this section, we evaluate LOOC on four dense scene datasets: UCSD~\cite{chan2008privacy}, Trancos~\cite{guerrero2015Trancos}, Mall~\cite{chen2012feature}, and Penguins~\cite{arteta2016counting}. For each of these datasets we only use the count labels instead of the original point-level annotations. For evaluation, we use  mean-absolute error (MAE) for measuring counting performance, and grid average mean absolute error (GAME)~\cite{guerrero2015Trancos} for localization performance.

For localization, we compare LOOC against a proposed baseline called TopK. The difference between TopK and LOOC is that TopK uses the fixed scores provided by the proposal generator to score the proposals and LOOC uses the dynamic scores provided by the object localizer's class probability map (CPM). 

We also compare LOOC against Glance, a state-of-the-art counting method that also uses count supervision. While Glance does not localize objects, the purpose of this benchmark is to observe whether the location awareness provided by LOOC can help in counting. Our models use the ResNet-50 backbone for feature extraction~\cite{he2016deep}, and they are optimized using ADAM~\cite{kingma2014adam} with a learning rate of 1e-5 and a weight decay of $0.0005$. We also got similar results using optimizers that do not require defining a learning rate~\cite{vaswani2019painless,loizou2020stochastic,vaswani2020adaptive}.

\textbf{UCSD}~\cite{chan2008privacy} consists of images collected from a video camera at a pedestrian walkway. This dataset is challenging due to the frequent occurrence of overlapping pedestrians, which makes counting and localization difficult. Following Li et al.~\cite{li2018csrnet}, we resize the frames to 952x632 pixels using bilinear interpolation to make them suitable for our ResNet based models. We use the frames 601-1400 as training set and the rest as test set, which is a common practice~\cite{chan2008privacy}, . 

\begin{figure}
  \centering
  \includegraphics[trim={1cm 2cm 1cm 1cm},clip, width=\columnwidth]{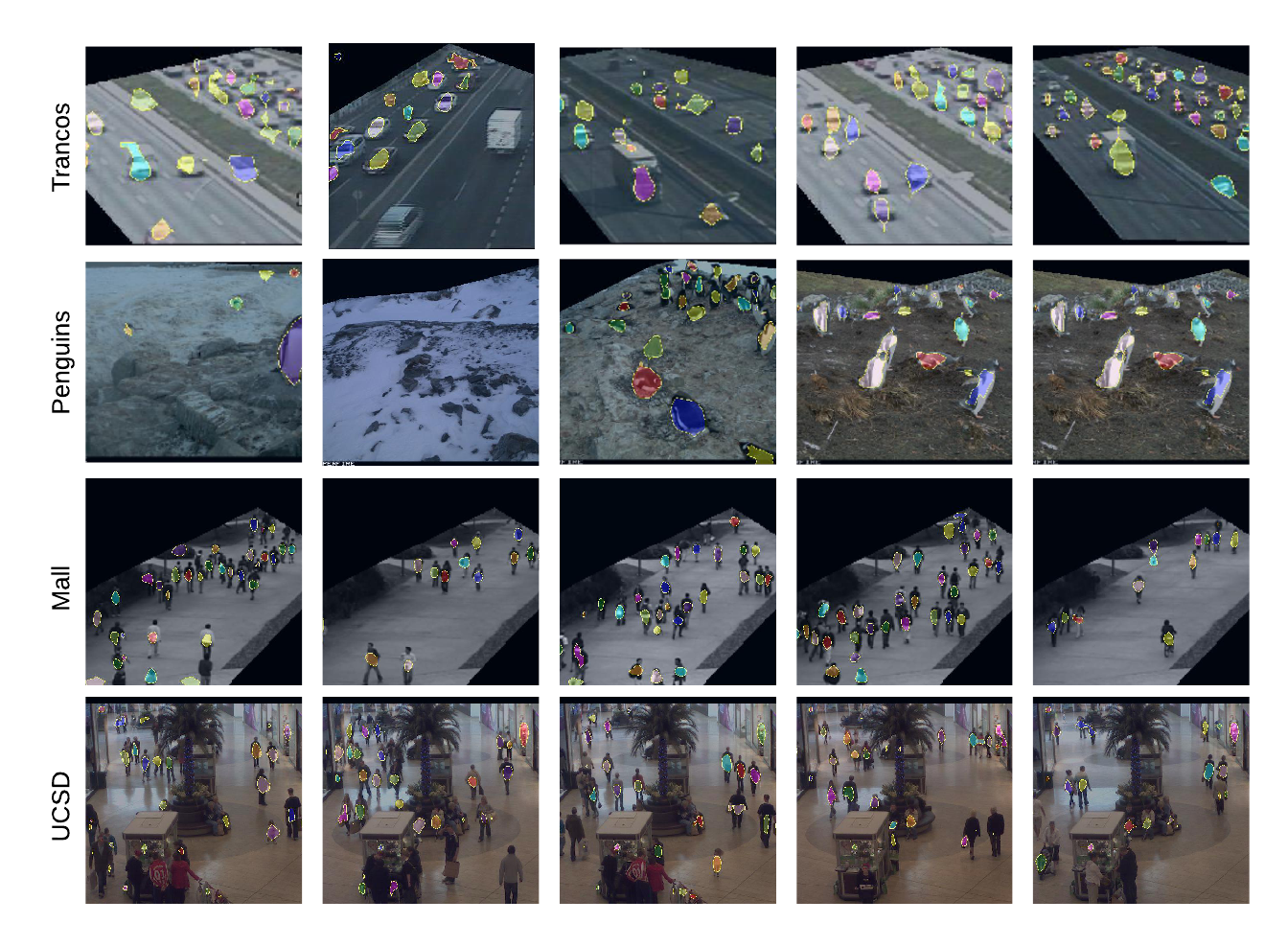}
  \caption{\textbf{LOOC predictions} on the 4 dense scene datasets.}
  \label{fig:qualitative}
\end{figure}

Table~\ref{tab:results} shows that LOOC outperforms Glance in terms of MAE, suggesting that localization awareness helps in counting as well. Further, LOOC outperforms TopK with respect to MAE and GAME suggesting that LCFCN provides informative class probability map. LOOC's results are also close to the fully supervised LCFCN, which indicates that good performance can be achieved with less costly labels. Qualitatively, LOOC is able to accurately identify pedestrians for UCSD (Figure~\ref{fig:qualitative}).

\textbf{Trancos}~\cite{guerrero2015Trancos} consists of images taken from traffic surveillance cameras for different roads, where the task is to count vehicles, which can highly overlap~\cite{guerrero2015Trancos}, making the dataset challenging for localization.

The results shown in Table~\ref{tab:results} indicate that LOOC achieves lower MAE than Glance, yet it can perform good localization compared to TopK. Compared to the fully supervised LCFCN, LOOC performs poorly mainly due to the quality of the pseudo point-level annotations, but the qualitative results appear accurate (Figure~\ref{fig:qualitative}).

\textbf{Mall}~\cite{chen2012feature}  consists of $2000$ frames of size $320\times240$ collected from a fixed camera installed in a shopping mall.  These frames have diverse illumination conditions and crowd densities, and the objects vary widely in size and appearance. The results in Table~\ref{tab:results} show that LOOC achieves good localization performance compared to TopK and counting performance compared to Glance.

\textbf{Penguins Dataset~\cite{arteta2016counting}}  consists of images of penguin colonies collected from fixed cameras in Antarctica.  We train on 500 images, and test on 500 unseen images. The quantitative results in Table~\ref{tab:results} and qualitative results in Figure~\ref{fig:qualitative} show the effectiveness of LOOC in scenes where objects can come in different shapes and sizes, and can densely overlap. 

\textbf{Ablation studies}. We evaluate the quality of the pseudo point-level annotations provided by LOOC in Table~\ref{tab:ablation_results}. After training LOOC, we generate the pseudo labels as the centroids of the top scoring $k$ proposals on the training set and measure the GAME localization score. We observe that LOOC outperforms TopK, suggesting that relying on LCFCN's class probability map allows us to score the proposals better. Thus, given count-level supervision, we can use LOOC to obtain high quality point-level annotations and then effectively train a fully-supervised localization on those point labels.

%%%%%%%%%%%%%%%%%%%%%%%%%%%%%%%%%%
% CONCLUSION
%%%%%%%%%%%%%%%%%%%%%%%%%%%%%%%%%%
\section{Conclusion}
 
We have proposed LOOC, a method that Localizes Overlapping Objects using Count supervision. LOOC trains by alternating between generating pseudo point-level annotations and training a fully supervised localization method such as LCFCN. The goal is to progressively improve the localization performance based on pseudo labels. The experiments show that LOOC achieves a strong new baseline in the novel problem setup of localizing objects using only count supervision. They also show that LOOC is a new state-of-the-art for counting in this weakly supervised setup. The experiments also show that the pseudo point-level annotations obtained by LOOC are of high quality and can be used to train any fully supervised localization method. For future work, we plan to investigate proposal free methods, perhaps those that rely on topological regularization~\cite{clough2019topological} to identify the regions of interest. Further, we also plan to look into incorporating regularization methods that help when the amount of labels is limited~\cite{verma2019manifold, rodriguez2020embedding}.

% divides the training images into labeled and unlabeled regions based on the scores of proposals generated by a common proposal method such as selective search. Using only the highest scoring proposals, this step generates a first set of pseudo point-level annotations. An object localizer such as LCFCN is trained on the labeled regions, which can then generate a class probability map (CPM) to identify the objects in the unlabeled regions. CPM is then used to re-score the proposals in order to obtain a new set of pseudo point-level annotations. This process is repeated until all the regions are labeled. Finally, a fully supervised object localizer is trained on the final set of pseudo point-level annotations.

%We have shown the efficacy of LOOC on four datasets dense scene datasets.  For counting, LOOC outperforms Glance which is a current state-of-the-art for counting that only uses count as its supervision, although it does not localize. The experiments also show that pseudo point-level annotations obtained by LOOC are high quality and can be used to train any fully supervised localization method.

%%%%%%%%%%%%%%%%%%%%%%%%%%%%%%%%%%
% BIBLIOGRAPHY
%%%%%%%%%%%%%%%%%%%%%%%%%%%%%%%%%%

%\bibliographystyle{IEEEbib}
\bibliography{refs}

\end{document}